\documentclass{article}

\usepackage{graphicx}

\usepackage{appendix}
\usepackage{amsmath}
\usepackage{algorithm}
\usepackage[noend]{algpseudocode}

\makeatletter
\def\BState{\State\hskip-\ALG@thistlm}
\makeatother

\usepackage[preprint]{neurips_2024}

\usepackage[utf8]{inputenc} 
\usepackage[T1]{fontenc}    
\usepackage{hyperref}       
\usepackage{url}            
\usepackage{booktabs}       
\usepackage{amsfonts}       
\usepackage{nicefrac}       
\usepackage{microtype}      
\usepackage{xcolor}         

\title{Diffusion based multi-domain neuroimaging harmonization method with preservation of anatomical details}

\author{
  Haoyu Lan\\
  University of Southern California\\
  Los Angeles, CA\\
  \texttt{haoyulan@usc.edu}\\ 
  \And 
  Bino A. Varghese\\
  University of Southern California\\
  Los Angeles, CA\\
  \AND 
  Nasim Sheikh-Bahaei\\
  University of Southern California\\
  Los Angeles, CA\\
  \And 
  Farshid Sepehrband\\
  University of Southern California\\
  Los Angeles, CA\\
  \And 
  Arthur W Toga\\
  University of Southern California\\
  Los Angeles, CA\\
  \And 
  Jeiran Choupan\\
  University of Southern California\\
  Los Angeles, CA\\
  NeuroScope Inc., NY\\
}

\begin{document}

\maketitle

\begin{abstract}
  Multi-center neuroimaging studies face technical variability due to batch differences across sites, which potentially hinders data aggregation and impacts study reliability. Recent efforts in neuroimaging harmonization have aimed to minimize these technical gaps and reduce technical variability across batches. While Generative Adversarial Networks (GAN) has been a prominent method for addressing image harmonization tasks, GAN-harmonized images suffer from artifacts or anatomical distortions. Given the advancements of denoising diffusion probabilistic model which produces high-fidelity images, we have assessed the efficacy of the diffusion model for neuroimaging harmonization. we have demonstrated the diffusion model’s superior capability in harmonizing images from multiple domains, while GAN-based methods are limited to harmonizing images between two domains per model. Our experiments highlight that the learned domain invariant anatomical condition reinforces the model to accurately preserve the anatomical details while differentiating batch differences at each diffusion step. Our proposed method has been tested on two public neuroimaging dataset ADNI1 and ABIDE II, yielding harmonization results with consistent anatomy preservation and superior FID score compared to the GAN-based methods. We have conducted multiple analysis including extensive quantitative and qualitative evaluations against the baseline models, ablation study showcasing the benefits of the learned conditions, and improvements in the consistency of perivascular spaces (PVS) segmentation through harmonization. 
\end{abstract}

\section{Introduction}

Magnetic resonance imaging (MRI) data acquired from different batches, including various imaging sites or scanners, has shown promise in improving sample sizes for neuroscience studies that rely on imaging and predictive efforts \cite{bethlehem2022brain} \cite{marek2022reproducible} \cite{van2014multi}. However, these neuroimaging data from multiple batches are susceptible to non-biological and technical variations that occur between data acquisition from different batches, commonly known as batch effects. These batch effects can arise from disparities in acquisition protocols, scanner manufacturers, scanner drift, and hardware imperfections \cite{jovicich2006reliability}. The existence of batch effects can potentially pose challenges in terms of reproducibility of neuroscience studies, generalizability of prediction algorithms, and effective integration of radiomics-based imaging biomarkers in clinical practice \cite{grech2015multi}. Neglecting to consider the acknowledged interference of batch effects can result in a reduction in effectiveness, less reliable discoveries, and potentially distorted results. Various types of methods have been investigated to harmonize the neuroimaging data to remove the batch effects ranging from statistical methods focusing on correcting the feature distributions \cite{fortin2017harmonization} \cite{bayer2022site} \cite{fortin2016removing} to image processing methods focusing on harmonizing the raw MRI images directly \cite{cackowski2023imunity} \cite{chang2022self} \cite{yao2022novel} \cite{zuo2021unsupervised} \cite{dinsdale2021deep} \cite{ren2021segmentation}. There are limitations commonly recognized in the harmonization tasks, such as, possibly introduced correlation between subjects during the harmonization process due to the entangled batch effects and biological effects. Therefore, the efficient separation between biological effects and batch effects and a precise estimation of these effects are imperative.

To address the harmonization problem in neuroimaging data and overcome the existing issues in the harmonization methods, we attempted to utilize the diffusion probabilistic model \cite{ho2020denoising} and our contributions are as follows: 

\begin{itemize}
\item We formulated the neuroimaging harmonization problem as a controllable domain adaptation problem. 
\item We proposed the skip sampling step to reduce the sampling steps for the harmonization purpose. 
\item We tested the proposed method in two public datasets ABIDE II and ADNI 1:
\begin{enumerate}
\item Our diffusion model-based harmonization method achieved superior harmonization results compared to GAN based models on ABIDE II dataset for site effects harmonization. We also visually demonstrated the effectiveness of multiple diffusion trajectories learning of the proposed method. 
\item We evaluated the scanner effects harmonization by downstream PVS segmentation analysis on ADNI 1 dataset and showed improved consistency of the PVS count ratio across scanners. 
\end{enumerate}
\end{itemize}

In this study, we show that denoising diffusion model could effectively address the batch effects harmonization with the preservation of anatomical details. We also found the learned condition improves the harmonization performance compared to fixed condition (edge map). In addition, diffusion model provides strong ability of learning multiple diffusion trajectories controlled by the domain embeddings, which enables the multi-domain harmonization through only one single model. Compared to GAN based harmonization method, diffusion model also has several other advantages such as  stable training and high-fidelity image generation. 

\begin{figure*}
  \centering
   \includegraphics[width=1\linewidth]{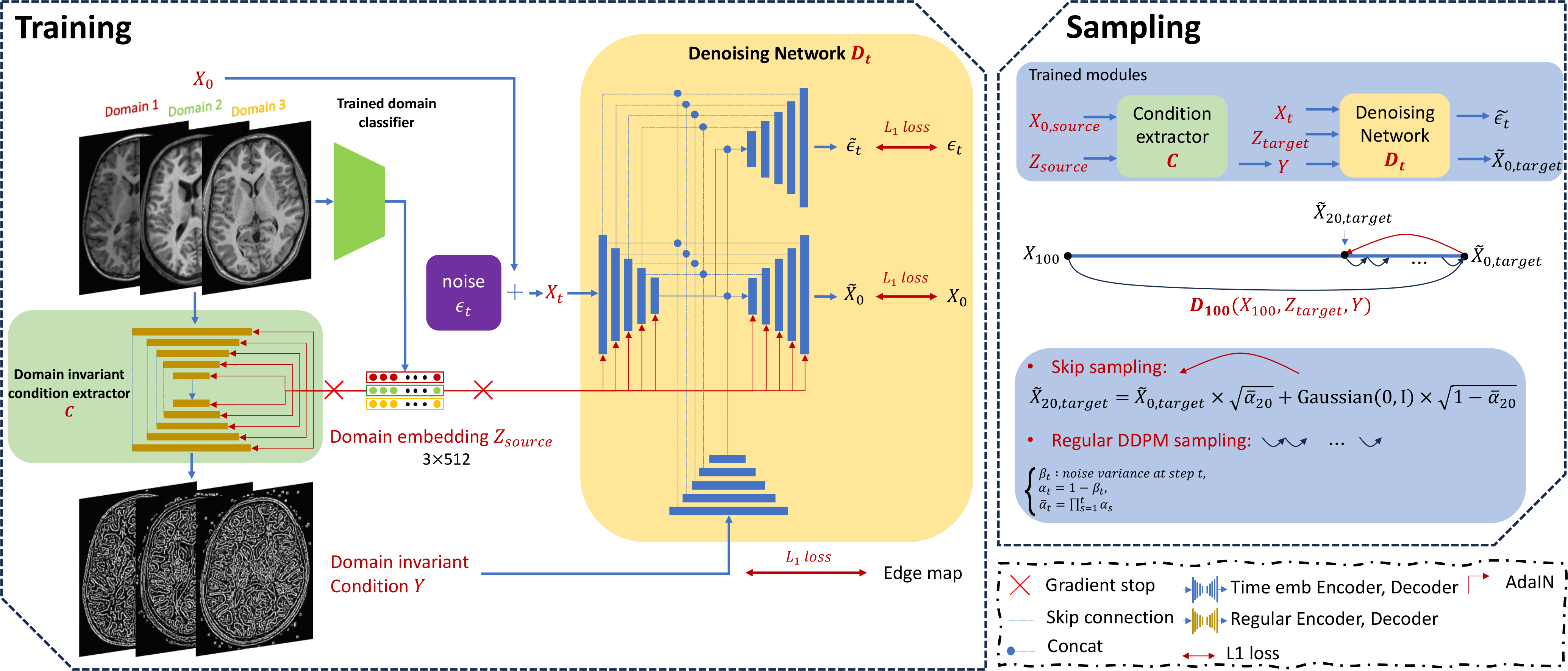}
   \caption{\textbf{The overall framework of the proposed method with training and sampling details.}}
\end{figure*}

\section{Related work}
The field of generative models has garnered significant interest in recent years due to their remarkable capability to learn the joint distribution. Mostly adopted models such as generative adversarial networks \cite{goodfellow2020generative}, variational autoencoder \cite{doersch2016tutorial}, and probabilistic diffusion model \cite{ho2020denoising} \cite{song2020score} also had a great impact in the medical image analysis field. GAN models have been popularly implemented in the neuroimage harmonization task. \cite{chang2022self} modeled CycleGAN \cite{zhu2017unpaired} to generate unpaired synthetic target scanner style image, then used the original image to perform histogram matching with the synthetic target scanner style image and achieved style harmonization. \cite{yao2022novel} proposed a style – content disentangle GAN to harmonize between the contrast enhanced T1w and high resolution T2w. Because batch effects are implicitly defined as style, this method has the risk of incorrectly including biological effects as the style. \cite{zuo2021unsupervised} proposed a variational autoencoder based model Contrast Anatomy Learning and Analysis for MR Intensity Translation and Integration (CALAMITI). CALAMITI encodes image into low-dimensional style-invariant content representation, then generates target batch image with content representation combined with style information. CALAMITI has been validated on multiple studies and showed its effectiveness, however, CALAMITI requires at least two MRI modalities to learn the batch effect invariant anatomy, which might not be doable in the clinical setting where T1w is the only available modality. \cite{ren2021segmentation} aimed to remedy the artifacts issue of CycleGAN by proposing a Segmentation-Renormalized (Seg-Renorm) framework with an additional semantic extractor of which the output was used to calculate the trainable scale and shift parameters to normalize each layer of the CycleGAN. However, most of the unpaired GAN based generative models only reciprocally translate two sets of domains of images per model. This means that, for the harmonization tasks that involve more than two domains, there will be a need to train more than one GAN models.

Considering the advantages of the diffusion model, including the stable training, high-fidelity image generation, and the potential learning multiple diffusion trajectories, we explored and evaluated the ability of diffusion model in the multi-domain neuroimaging harmonization and proposed a denoising diffusion model-based harmonization framework.

\section{Proposed Method}
\subsection{Multi-domain neuroimaging harmonization as a controllable domain adaption problem }
Neuroimaging data acquired at different manufactured scanners, or at different imaging sites often contain non-biological domain shifts introduced by these technical variabilities \cite{hu2023image}\cite{bayer2022site}. We approached the multi-domain neuroimaging harmonization as a controllable domain adaption task \cite{farahani2021brief}, which aims to learn the batch effects by a domain embedding controlled denoising diffusion process. Since one of the remained challenges in the most neuroimage harmonization techniques is the lack of the ability to disentangle the biological effects and batch effects during the harmonization process, we conditioned the diffusion process with a learned domain invariant condition. This condition reinforces preservation of the anatomical details during the diffusion harmonization process. 

We introduced a denoising diffusion model based multi-domain harmonization framework (Figure 1) which includes a domain invariant condition extractor $C$ and a denoising network $D_{t}$. Both modules were controlled by the domain embedding which determines the diffusion trajectory. 

\begin{figure*}
  \centering
   \includegraphics[width=0.75\linewidth]{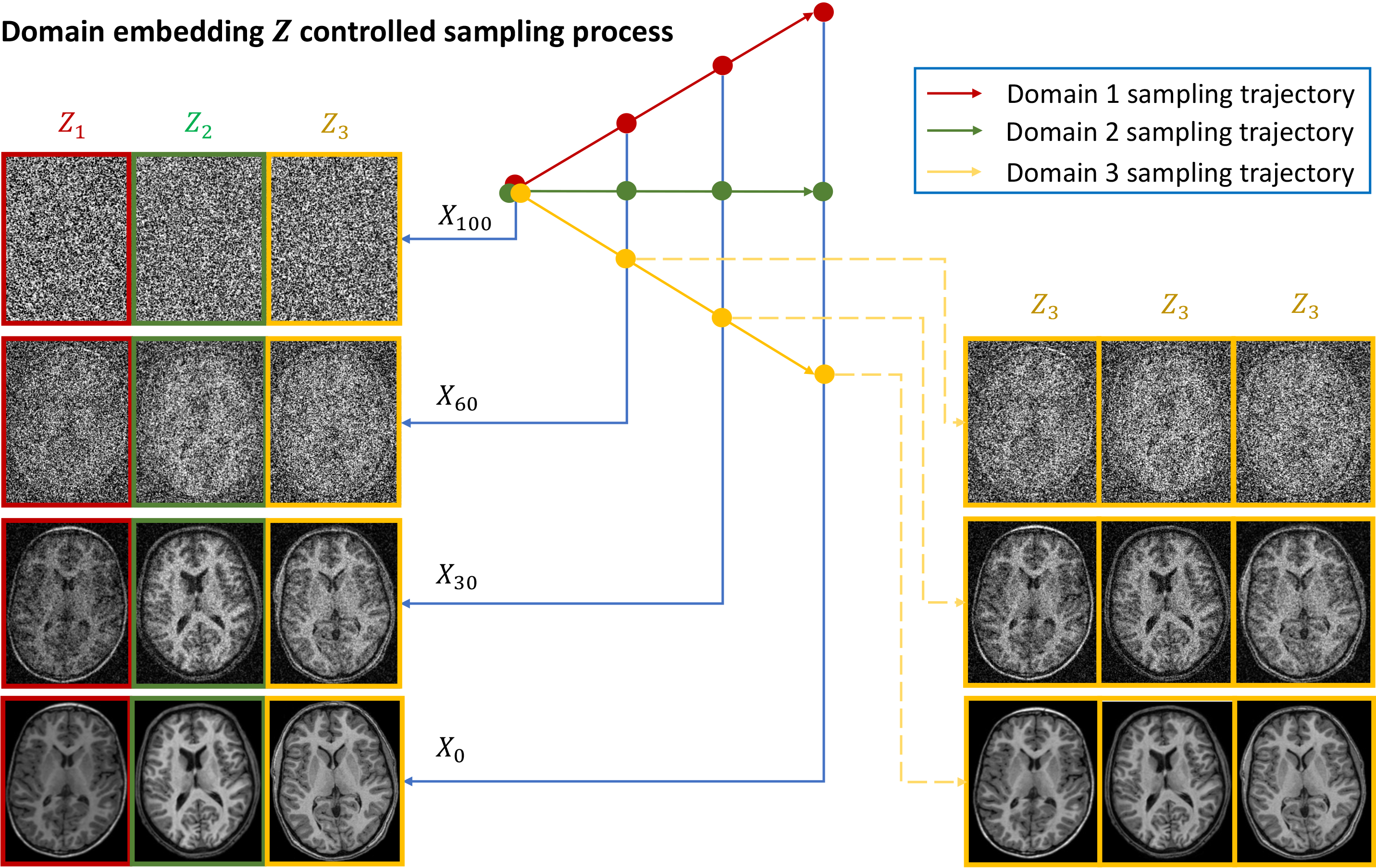}
   \caption{\textbf{Visual demonstration of the learned trajectories.} Domain embedding Z steers the sampling trajectories to achieve the multi-domain harmonization. Comparison between left and right side of the figure demonstrates the effectiveness of harmonization at each sampling step. }
\end{figure*}

\subsection{Learned domain invariant conditioned controllable denoising diffusion model}
 Denoising diffusion probabilistic model (DDPM) learns to capture the underlying structure of clean images and the noise characteristics at each time step, enabling DDPM to effectively denoise new, unseen images. The denoising network $D_{t}$ was implemented following DDPM paper which is a positional time embedding t controlled U-Net \cite{ronneberger2015u} model with slight modifications. $D_{t}$ has two encoders and two decoders. As shown in Figure 1, the encoder receiving the noise corrupted image $X_{t}$ is controlled by the domain embedding, while the encoder taking the domain invariant condition Y is not controlled; the decoder predicting the uncorrupted image $X_{0}$ is controlled by the domain embedding, while the decoder estimating the standard gaussian noise $\tilde\epsilon_{t}$ is not controlled. The rationale of this design is due to domain invariant nature of $Y$ and $\tilde\epsilon_{t}$ . 

The domain invariant condition extractor C is a regular U-Net model which are controlled by the domain embedding and is regarded as a function to project images from different image domains to the same domain invariant condition domain. To correlate the domain embedding with the denoising network and the condition extractor, we utilized Adaptive Instance Normalization (AdaIN )\cite{huang2017arbitrary}  to normalize each layer in the target layers of the modules. Given the domain invariant condition \(Y = {\bf C}(X_{0}, Z_{source})\) and denoising outuputs \(\tilde\epsilon_{t}, \tilde X_{0} = {\bf D_{t}}(X_{t}, Z_{source}, Y)\), the loss functions of the proposed method is as follow: 

\begin{equation}
    L_{noise} = |\tilde\epsilon_{t}-\epsilon_{t}|, L_{recon} = |\tilde X_{0}-X_{0}|, L_{cond} = |Y-canny\,edge|
\end{equation}

where $Z_{source}$ is the domain embedding of $\tilde X_{0}$ and \(L_{Total} = L_{noise}+L_{recon}+L_{cond}\). 

As shown in Figure 1, $L_{noise}$  and $L_{recon}$ update the denoising network $\bf D_{t}$, while all three components of the $L_{total}$ update the domain invariant condition extractor $\bf C$. Three components of the loss function reinforce the condition extractor to extract the condition which is domain invariant and is imbued with enough anatomical details for the domain embedding controlled denoising network to estimate the noise at step t and predict the original input data. For our experiments which focus on imaging texture harmonization, canny edge detector generated edge map was a good ground truth to calculate $L_{cond}$.

We trained an independent image domain classifier and used its layer with 512 elements before the classification head as the domain embedding $Z_{source}$ (Figure 1). Since domain classifier is separated from the overall framework, training won’t update the classifier. 

\subsection{Skip sampling}
Assume we have 100 steps for the diffusion process and at the end of the diffusion, $X_{100}$ have been efficiently corrupted to the random noise. For the denoising network $\bf D_{100}$ to predict the uncorrupted images $X_{0}$, the only resource it can rely on is the domain invariant condition $Y$ and domain embedding $Z_{target}$. This feature makes the domain style prediction of $X_{0}$ at diffusion step 100 purely controlled by $Z_{target}$ and anatomical prediction of $X_{0}$ purely conditioned on $Y$. One point to note, $Z_{source}$ was used for the training of $D_{t}$ during the training process (as shown in Figure 1), however, at the sampling step $Z$ could be any target domain embedding.

The key to the sampling step (pseudocode could be found in Appendix A) for the harmonization is that the target domain $\tilde X_{0,target}$ is predicted at step 100, then we corrupt the $\tilde X_{0,target}$ back to $\tilde X_{20,target}$. For this step we call it skip sampling.  After the skip sampling, we use regular DDPM sampling to sample the image from $\tilde X_{20,target}$ to $\tilde X_{0,target}$. Since direct prediction of $\tilde X_{0,target}$ from $X_{100}$ still misses details of target domain style, by doing skip sampling we could shorten the DDPM sampling process and generate images with rich target domain style as shown in Figure 3. Selection of the step for the skip sampling is based on the empirical study of the given dataset. 

\begin{figure*}
  \centering
   \includegraphics[width=1\linewidth]{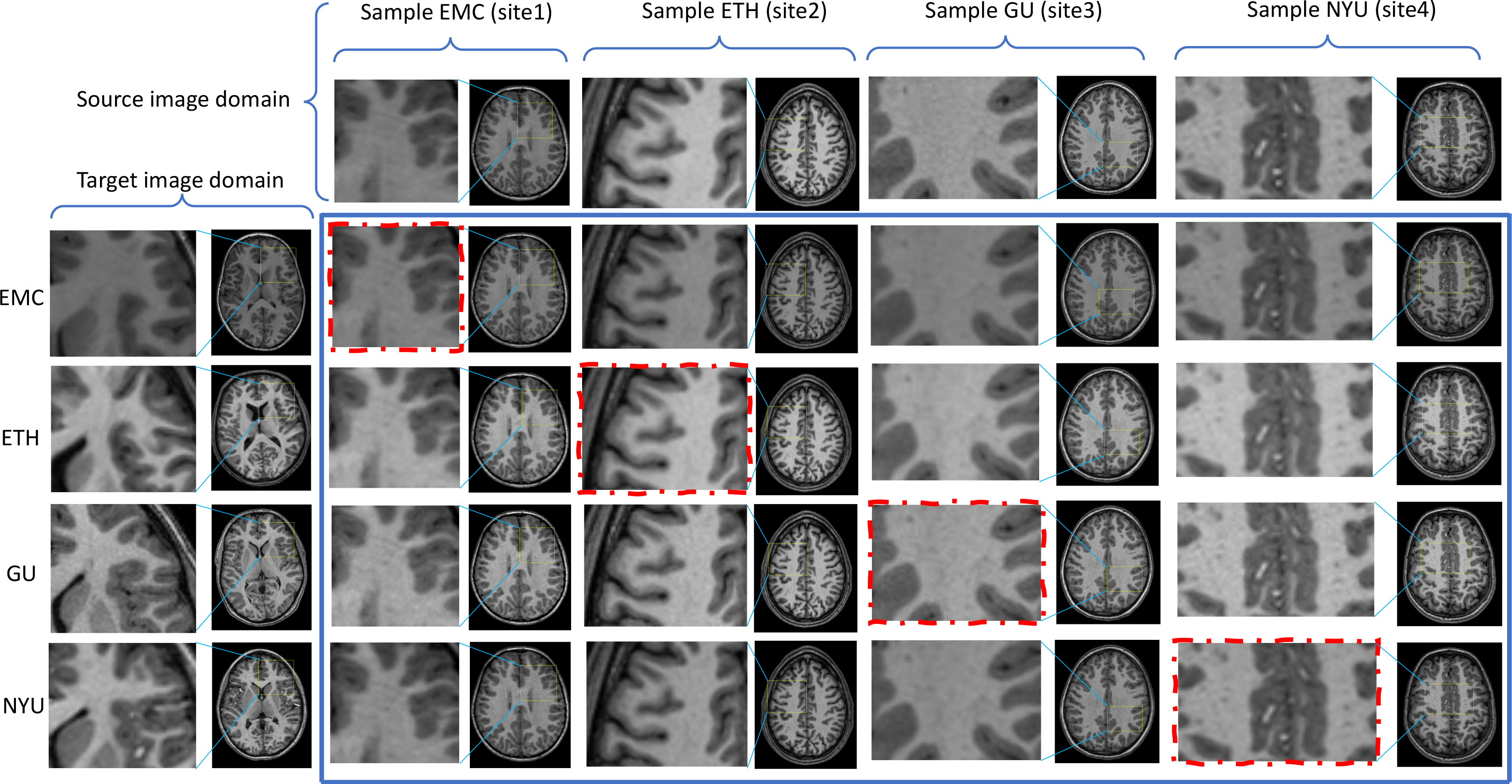}
   \caption{\textbf{Qualitative results of the harmonization on ABIDE II.} The far-left column shows samples of target image domains (reference of imaging texture), and the top row shows samples of source image domain (reference of anatomical details). The red dash squares highlight the harmonization case where target image domain equals to the source image domain. }
\end{figure*}

\section{Experiments and Results}
\subsection{Dtaset}
$\bf ADNI\,1:$ We selected the older participants of the Alzheimer’s Disease Neuroimaging Initiative (ADNI 1) \cite{jack2008alzheimer} cohort as one of the experiment datasets. 456 participants, aged between 55 and 90 years were gathered during the 12-month follow-up period after recruitment. All 456 subjects’ MRI images collected for this study follow the same image acquisition: T1w MPRAGE MRI images using scanners manufactured by either GE Healthcare, Philips Medical Systems, or Siemens Medical Solutions operating at 1.5T field strength. These images had a voxel resolution of $1.25\times1.25\times1.2 mm^3$ and were obtained with the following acquisition parameters: TR=2400 ms, TE= 3.6 ms, flip angle=8.0 degrees, FOV=24 cm, and slice thickness=1.2mm.

$\bf ABIDE\,II:$ We selected data of four sites from the Autism Brain Imaging Data Exchange (ABIDE II) \cite{di2014autism} as one of the experiment datasets including 48 subjects from Erasmus University Medical Center Rotterdam (EMC), 36 subjects from ETH Zurich (ETH), 90 subjects from Georgetown University and 78 subjects from NYU Langone Medical Center (NYU). Detailed acquisition parameters could be found in Table 2.

we selected axial slices from each subject with brain signals (whole background signal slices were discarded) which resulted in 45600 training samples/11400 testing samples for ADNI 1; 35280 training samples/8820 testing samples for ABIDE II. Detailed image preprocessing steps could be found in Appendix C.

\subsection{Perivascular spaces}
PVS \cite{wardlaw2020perivascular}\cite{barisano2022imaging}, also known as Virchow-Robin spaces, are fluid-filled spaces that surround penetrating blood vessels in the brain. These spaces are lined by a layer of cells and contain cerebrospinal fluid (CSF). PVS play an important role in the brain waste clearance exchanging nutrients and waste products between CSF and brain tissue. They are more visible in T2-weighted MRI sequences, in comparison with T1-weighted sequences. In certain conditions or diseases, PVS may become enlarged or show abnormalities, which can be indicative of underlying health issues. Frangi filter \cite{frangi1998multiscale} is an image processing algorithm that detects tubular structures in the 2D or 3D image spaces and commonly adopted to segment PVS in the T1w or T2w MRI images \cite{sepehrband2019image}\cite{lan2023weakly}\cite{ballerini2018perivascular}. To generate the PVS count, binarized PVS map would be generated at first. Then the regions with less than 3 connected voxels would be removed, which could be potential random noise or motion corruption captured by frangi filter. Finally, the overall PVS count would be divided by total white matter volume to generate the PVS count ratio as a more generalized measure. We used PVS count ratio as a evaluation metric of scanner effects existing in ADNI 1. 

Batch effects in the T1w images from ADNI 1 are the scanner effects and we focused on the PVS segmentation as the downstream analysis to demonstrate the harmonization efficacy. Batch effects in the T1w images from ABIDE II are the site effects and we reported FID score \cite{heusel2017gans} and qualitative visualization to demonstrate the harmonization efficacy.

\subsection{Experimental setup}
The denoising network was designed following original DDPM U-Net noise estimation model with 5 skip-connection levels and each level was constituted by two ResNet \cite{he2016deep} blocks each with two convolutional layers. Self-attention \cite{vaswani2017attention} was calculated at down-sampling and up-sampling path at third level and bottleneck layers. The condition extractor was designed the same way as the denoising network with only one encoder, one decoder, and not controlled by the positional time embedding. Noise variance $\beta$ was quadratically schedule between 0.0001 and 0.5 with 100 steps as a fixed sequence. CycleGAN and Seg-Renorm models were downloaded from their public GitHub repositories and finetuned accordingly with our experimental datasets. All experiments were implemented and conducted using Python 3.8 and PyTorch 1.13.1 and deployed on NVIDIA GPU cluster equipped with eight V100 GPUs for training and testing. 

The separately trained domain classifier was a ResNet with 6 levels and 2 convolution layers at each level. Self-attention was calculated before the sixth level. The domain classifier has been trained to classify the corresponding image domain and we used its multilayer perceptron layer with 512 elements before the classification head as the domain embedding. All models were trained and reached optimal results on the validation data. 

\begin{figure*}
  \centering
   \includegraphics[width=0.75\linewidth]{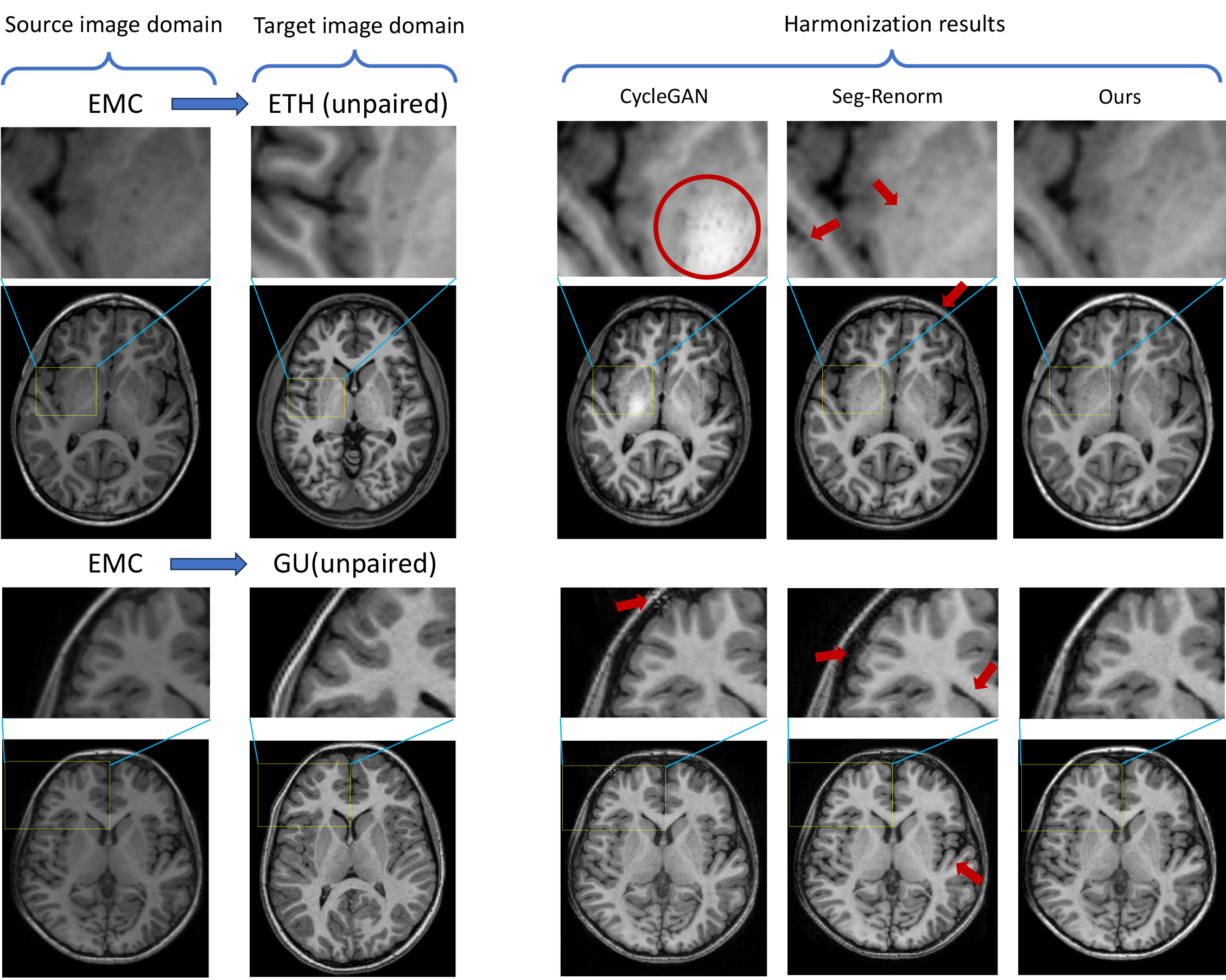}
   \caption{\textbf{Qualitative comparison with baseline models.} Red arrows indicate the areas with generated artifacts and mild stripe patterns which are not present in both source and target image domain. }
\end{figure*}

\subsection{Domain embedding controls DDPM to learn multiple diffusion trajectories}
Domain embedding controls diffusion model to learn multiple diffusion trajectories at once which enables multi-domain harmonization with single model. Figure 2 is the qualitative visualization of the diffusion sampling process for the ABIDE II dataset. One sample of each of the three sites was randomly selected for this visualization. To demonstrate multiple intermittent steps of the diffusion process of the harmonization task, skip sampling was not used. $Z_1$ (red), $Z_2$ (green) and $Z_3$ (yellow) represent the three domain embeddings generated by the pretrained domain classifier and colored arrows represent different sampling trajectories. Samples from four time-steps $X_{100}$,$X_{60}$,$X_{30}$,$X_{0}$ are shown for visual comparison. For the images on the left side, $Z_{target}$ equals to $Z_{source}$; for the images on the right side, $Z_{target}$ is $Z_3$. As time T goes from 100 to 0, the site difference (approximately indicated by the distance between each pair of dots at each time step) is increasing, which can also be visually verified by image samples’ differences at each time step on the left side. This indicates that the denoising network learned multiple diffusion trajectories and correctly correlated to the control domain embedding. By controlling sampling process with $Z_3$ on the right side, the image samples' differences at each time step were reduced resulting in harmonized images.

\subsection{Evaluation of harmonization on ABIDE\,II }
Figure 3 shows the harmonization results of image samples from four sites EMC, ETH, GU and NYU of ABIDE II to the target image domains respectively. The far-left column shows samples of target image domains (reference of imaging texture), the top row shows samples of source image domain (reference of anatomical details), and the rest of the images show harmonization results to the corresponding target image domain. The red dash squares highlight the harmonization case where target image domain equals to the source image domain. Within the zoomed in regions, it could be easily noted that the source images have been harmonized to the target image domain with corresponding textures and contrasts and have anatomical details preserved.

\begin{table}[!h]
  \caption{\textbf{Quantitative comparison with baseline models and the ablation model.} The \textbf{Bold} font indicates the best and \underline{underline} indicates the second best. The qualitative comparison can be found in Figure 4 and Figure 6. }
  
  \centering
  \resizebox{\columnwidth}{!}
  {
  \begin{tabular}{lllll}
    \toprule& \multicolumn{4}{c}{FID $\downarrow$}\\
    \cmidrule(r){2-5}
    Method     & Others $\rightarrow$ EMC     & Others $\rightarrow$ ETH  & Others $\rightarrow$ GU & Others $\rightarrow$ NYU\\
    \midrule
    Source, Target & 39.6 & 56.2 & 36.7 & 37.6\\
    CycleGAN (baseline) & \bf{23.1}  & 40.1 & 18.4 & 31.1\\
    Seg-Renorm (baseline) & \underline{25.9} & \underline{39.4} & \underline{15.2} & 28.9\\
    Ours (Edge map) & 36.5 & 46.8 & 30.6 & \underline{25.2}\\
    Ours (Learned condition) & \bf{23.1} & \bf{38.9} & \bf{12.1} & \bf{20.5}\\
    \bottomrule
  \end{tabular}
  }
\end{table}

In order to quantitatively evaluate our proposed method and compare with other state-of-the-art GAN based unpaired harmonization methods, we calculated the FID scores and reported in Table 1. We evaluated two baseline models CycleGAN and Seg-Renorm on ABIDE II with four sites data included. Since both CycleGAN and Seg-Renorm harmonize two sites’ data per model, we trained 6 models for CycleGAN and 6 models for Seg-Renorm to harmonize all four sites’ data correspondingly. Similar to the original Seg-Renorm paper, FreeSurfer 7.1.1 \cite{fischl2012freesurfer} was used to generate the volumetric segmentation mask for the training purpose of Seg-Renorm. We reported four sets of FID scores with target image domain versus other three image domains as shown in Table 1, and InceptionV3 \cite{szegedy2016rethinking} was used as the feature extractor to calculate the FID score instead of a customized feature extractor to avoid any dataset induced biases. All three harmonization methods have FID scores reduced compared to the FID scores among the original images, which means that after the harmonization, harmonized images are perceptually more similar as the images from the target image domain. Proposed method had the best FID score for three harmonization cases and tied with CycleGAN at the harmonization case of the EMC site. 

Figure 4 shows qualitative evaluation of the harmonization results from all three methods. First two columns show samples from source image domain (reference of anatomical details) and unpaired sample from target image domain (reference of imaging texture). We noticed that CycleGAN has universal artifacts present in the generated images and might not be able to be correctly quantified by the FID score. Seg-Renorm was able to solve the universal artifacts issue in the CycleGAN generated images, however, it still contains mild stripe patterns which are not present in the source images. Additionally, there are mild anatomy distortion presented in the Seg-Renorm results (indicated by the red arrows). Proposed method generated the most anatomical details accurate and imaging texture aligned harmonization results. 

\begin{figure*}
  \centering
   \includegraphics[width=1\linewidth]{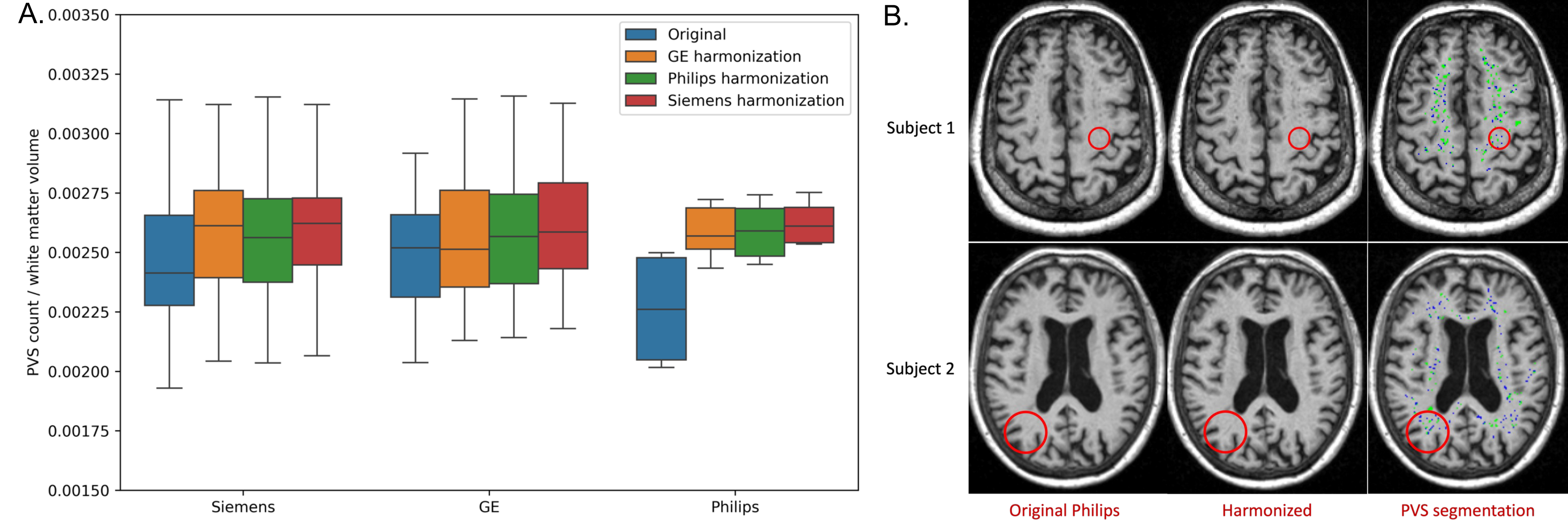}
   \caption{\textbf{PVS segmentation quantification.} \textbf{A.} Harmonized images show reduced heterogeneity regarding PVS count ratio across scanners as shown in orange, green and red boxplots compared to the blue boxplots. \textbf{B.} Visualization of the increased PVS count ratio derived from the harmonized image (blue PVS) in comparison with the original image (green PVS)}
\end{figure*}

\subsection{Harmonization improves consistency of PVS count ratio across MRI scanners on ADNI\,1}

As shown in Figure 5 A., we observed inconsistent Frangi filter segmented PVS count ratio distributions among images in ADNI 1, with lower PVS count ratio in the images acquired by Philips compared to images acquired from other two types of scanners. After the harmonization, PVS count ratio distribution became consistent across images acquired from all three scanners. Figure 5 B. shows the original image acquired at Philips scanner, the image harmonized to GE scanner style and PVS segmentation overlays. The red circles indicate regions with increased PVS contrast in harmonized images compared to original images. The blue overlay is the PVS segmentation from the harmonized image and the green overlay is the PVS segmentation from the original image. Note that since green overlay was stacked on top of blue overlay, visible blue voxels indicate the PVS which were missed by the segmentation from the original image. The same Frangi filter threshold was applied to both segmentations.

We used trained domain classifier to extract the domain embeddings from images for the t-SNE \cite{van2008visualizing} visualization which could be found in Figure 7. As shown in the Figure 7, t-SNE visualizes the extracted domain embedding vectors of test data in the two-dimensional space for ADNI 1 dataset. Test images in the original image space have formed three distinct clusters as shown in the top left corner of the figure. After the images get harmonized to either GE, Philips, or Siemens scanner effects, the separations disappear and all test data belong to one big cluster, which indicates that the harmonized images only contain one of the three scanner effects.

\begin{figure*}
  \centering
   \includegraphics[width=.75\linewidth]{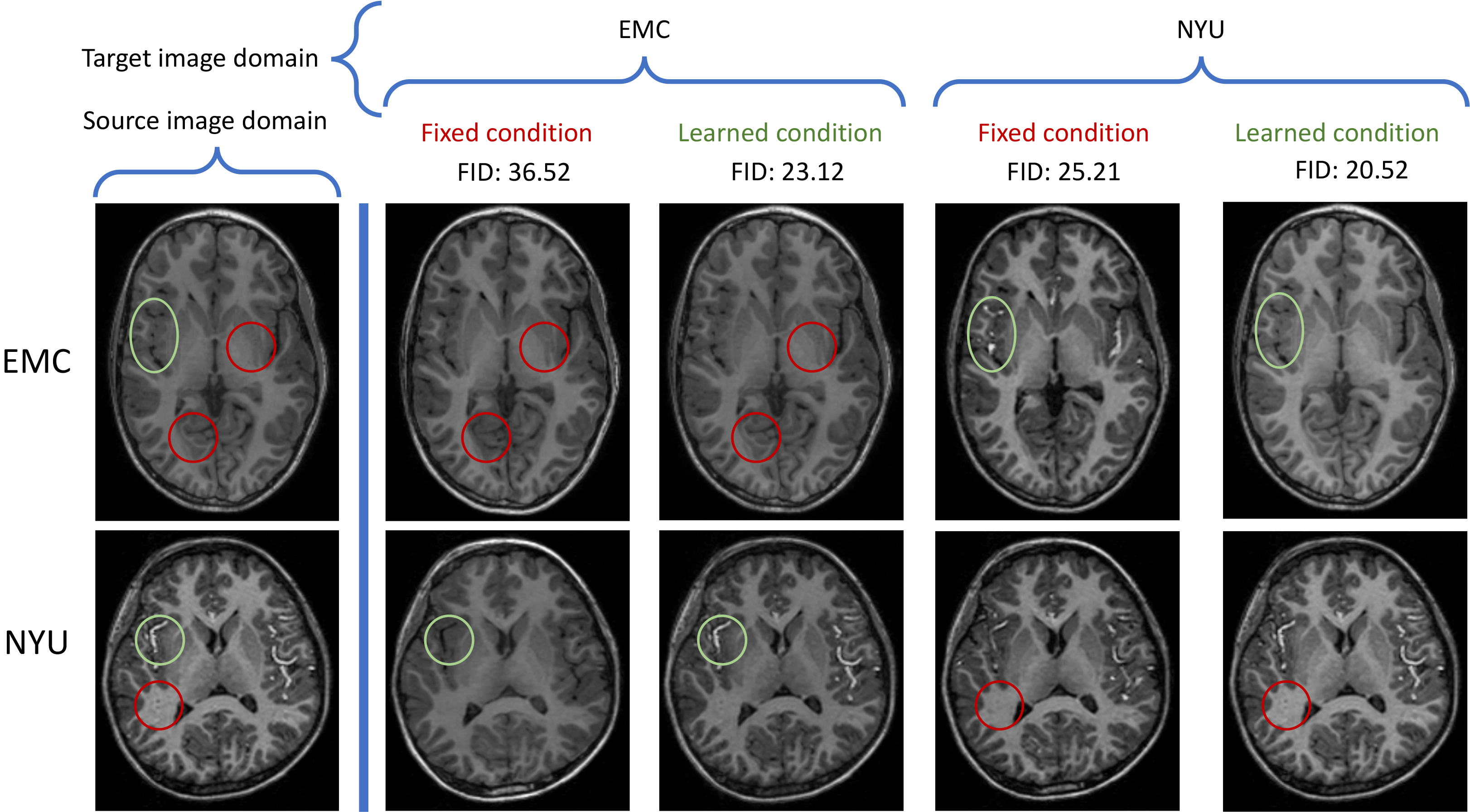}
   \caption{\textbf{Qualitative comparison of harmonization results with fixed condition and learned condition.} Green circles indicate that the artery signals (only sensitive in the NYU image domain) got removed by harmonizing NYU → EMC data domain and generated by harmonizing EMC →NYU data domain with fixed condition (edge map), while learned condition reinforced consistent artery signals between the source image and the harmonized target image.  Red circles indicate the areas with inferior performance of the preservation of anatomical details by conditioning on the fixed condition compared to the learned condition. }
\end{figure*}

\subsection{Ablation study (learned condition improves the generated image quality)}
We study alternative denoising network conditioning by removing the condition extractor network and instead using the edge map as the condition Y. As shown in Figure 6, samples from two image domains were presented and ‘Fixed condition’ represents the edge map conditioning. NYU data is the only site among four sites from ABIDE II that is sensitive to the artery signals. One interesting result from the edge map conditioning is the removed artery signals by harmonizing NYU data to EMC data domain and generated artery signals by harmonizing EMC data to NYU data domain as indicated by green circles. However, the anatomical details were not fully preserved after the harmonization with edge map conditioning as indicated by red circles.  In Table 1 we also reported inferior FID score of the proposed method with fixed condition compared to the learned condition.  

\section{Discussion and Conclusion}
We proposed a diffusion-based multi-domain harmonization framework which learns multiple diffusion trajectories correlated to image domains respectively. Diffusion model was conditioned on the learned domain invariant condition which guarantees the anatomical details unchanged during the harmonization. A skip sampling was introduced to shorten the sampling process. We quantitatively evaluated our proposed method on ABIDE II dataset by FID score for site effects harmonization and qualitatively demonstrated the superior performance of our method compared to two baseline GAN based models. Further, we evaluated our method on ADNI 1 dataset for scanner effects harmonization with an additional PVS segmentation downstream analysis to show improved consistency of PVS count ratio after the harmonization. 

\subsection{Limitations}
ADNI\,1 dataset used for the experiment was acquired under the 1.5T field strength with three imaging vendors; ABIDE II used for the experiment was acquired under 3T field strength from four imaging sites. For our experiment, the learned domain condition edge map for the diffusion model has been only tested for the images acquired under the same field strength and harmonization focused on the imaging texture heterogeneities. In this study, we did not perform the harmonization tasks on images acquired under different field strengths or incorporating the super-resolution, nor the reliability of learned domain condition edge map has been tested in those cases. Also, proposed method was only tested on the neuroimaging data, which is the focus of this study. We will be looking forward to testing the proposed method on more broad range of imaging datasets and other conditioning methods. 

\subsection{Conclusion}
In conclusion, our work showed efficacy of using diffusion model to tackle neuroimaging harmonization problem with the preservation of anatomical and biological details. Our work is specifically evaluated to harmonize the imaging texture heterogeneity caused by the technical variability present in the large cohorts of multi-center dataset. Such datasets are valuable to be used for investigating the association between clinical pathologies and differentiating their lasting impacts on the clinical manifestation of neurological diseases. For future work, it is recommended to explore a more efficient domain embedding method which is capable of capturing the hierarchical domain variant.


\bibliographystyle{plainnat}
\bibliography{main}

\newpage

\appendix
\appendixpage
\addappheadtotoc

\section{Skip sampling pseudocode }

\begin{algorithm}
\caption{}\label{euclid}
\begin{algorithmic}[1]
\State $\textit{$X_{0,source}$} \gets \text{Input image}$, $\textit{$Z_{source}$} \gets \text{Source domain emb}$, $\textit{$Z_{target}$} \gets \text{Target domain emb}$
\State $\textit{$X_{100}$} \gets \text{Gaussian(0,I)}$
\State $Y=\bf{C}(X_{0,source}, Z_{source})$
\State $\tilde X_{0,target}, \tilde \epsilon_{100}=\bf{D_{100}}(X_{100}, Z_{target},Y)$
\State $t=20$        
\State $X_{t}=\tilde X_{0,target}\times\sqrt{\tilde\alpha_{t}}+Gaussian(0,I)\times\sqrt{1-\overline\alpha_{t}}$ \quad\quad\quad\quad\quad\quad\quad\quad \textbf{Skip sampling}

\BState \emph{loop}:
\If {$t>0$}
\State $z \gets \text{Gaussian(0,I) \text{if} $t$>1, \text{else} $z$=0}$
\State $\tilde X_{0,target}, \tilde \epsilon_{t}=\bf{D_{t}}(X_{t}, Z_{target},Y)$
\State $X_{t-1}=\dfrac{1}{\sqrt{\alpha_{t}}}(X_{t}-\dfrac{1-\alpha_t}{\sqrt{1-\overline\alpha_{t}}}\times \tilde \epsilon_{t})+0.5\times log(1-\alpha_{t})\times z$ \quad\quad\quad \textbf{DDPM sampling}
\State $t=t-1$
\EndIf
\State \textbf{Return} $X_{0}$.
\end{algorithmic}
\end{algorithm}

\section{Acquisition parameters of selected sites from ABIDE\,II}

\begin{table}[!h]
  \caption{\textbf{Acquisition parameters of selected sites from ABIDE\,II.}}
  \centering
  \begin{tabular}{lllll}
    \cmidrule(r){2-5}
         & EMC     & ETH  & GU & NYU\\
    \midrule
    Field Strength & 3T & 3T &3T&3T\\
    Manufacturer & GE&Philips&Siemens&Siemens\\
    Model & MR750&Achieva&TriTim&Allegra\\
    Headcoil & 8ch&32ch&12ch&8ch\\
    Sequence  & IR-FSPGR&3D TFE&MPRAGE&3D TFL \\
    Slice Thickness [mm] & 0.9&0.9&1&1.33\\
    Slice In-Place Resolution [mm2] &0.9$\times$ 0.9&0.9 $\times$ 0.9&1$\times$ 1&1.3 $\times$ 1.0\\
    \bottomrule
  \end{tabular}
\end{table}

\section{Image preprocessing of the ANDI 1 and ABIDE II datasets}
FreeSurfer version 7.1.1 recon-all module on the Laboratory of Neuro Imaging (LONI) pipeline system (https://pipeline.loni.usc.edu) initially resampled all images to $256\times256\times256$ image size, $1\times1\times1$ mm3 voxel resolution, rescaled intensity range to $0-255$, and reoriented the images to RAS (Right, Anterior, Superior) coordinate system. Non-parametric Non-uniform intensity Normalization (N3) \cite{sled1998nonparametric} was used to correct the bias field through FreeSurfer. The bias filed correction using N3 was only meant to correct the bias field present equally in images acquired from different scanners or imaging sites. Preprocessed images have voxel intensity rescaled to $0-255$, which removed the intensity value range batch effect observed from images acquired from different sites and scanners.

\section{t-SNE visualization of harmonization effect on ANDI 1}
t-SNE visualization is shown in Figure 7. 
\begin{figure}
  \centering
   \includegraphics[width=.75\linewidth]{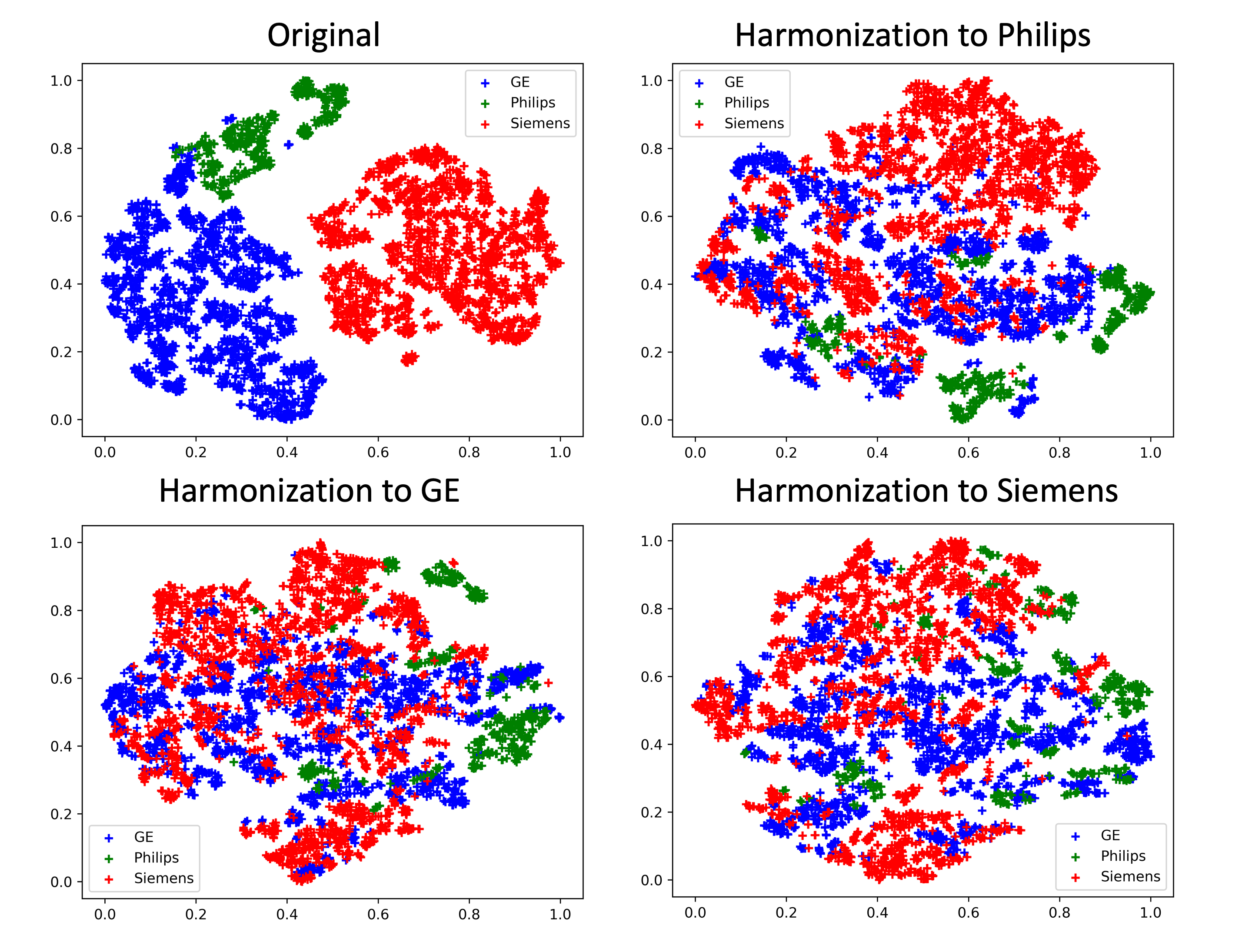}
   \caption{\textbf{t-SNE visualization of harmonization effect.} In the original image space, there are identifiable three separated clusters visualized by t-SNE plot as shown in the top left corner. After the images get harmonized to either GE, Philips, or Siemens image domain, the separations disappear and all test data belong to one big cluster, which indicates the effectiveness of scanner effect harmonization. t-SNE was calculated on the testing data.}
\end{figure}

\end{document}